\documentclass{article}
\usepackage{spconf,amsmath,graphicx,amssymb}
\usepackage{lineno}
\usepackage{hyperref}
\usepackage{enumitem}
\usepackage[font=small,skip=0pt]{caption}

\usepackage{algorithm}
\usepackage{algpseudocode}
\algnewcommand\algorithmicforeach{\textbf{for each}}
\algdef{S}[FOR]{ForEach}[1]{\algorithmicforeach\ #1\ \algorithmicdo}

\title{SCENE TEXT RECOGNITION MODELS EXPLAINABILITY USING LOCAL FEATURES}
\name{Mark Vincent Ty\textsuperscript{1}, Rowel Atienza\textsuperscript{1,2}}
\address{Electrical and Electronics Engineering Institute\textsuperscript{1} and AI Graduate Program\textsuperscript{2}, University of the Philippines \\
\{mark.vincent.ty, rowel\}@eee.upd.edu.ph}

\begin{document}
%
\maketitle
\begin{abstract}
Explainable AI (XAI) is the study on how humans can be able to understand the cause of a model’s prediction. In this work, the problem of interest is Scene Text Recognition (STR) Explainability, using XAI to understand the cause of an STR model’s prediction. Recent XAI literatures on STR only provide a simple analysis and do not fully explore other XAI methods. In this study, we specifically work on data explainability frameworks, called attribution-based methods, that explain the important parts of an input data in deep learning models. However, integrating them into STR produces inconsistent and ineffective explanations, because they only explain the model in the global context. To solve this problem, we propose a new method, STRExp, to take into consideration the local explanations, i.e. the individual character prediction explanations. This is then benchmarked across different attribution-based methods on different STR datasets and evaluated across different STR models.
\end{abstract}
\section{Introduction}
Scene Text Recognition (STR) \cite{baek2019wrong,yu2020towards,chen2021text} refers to the act of reading text from a natural scene setting. STR images are usually more difficult to predict than optical character recognition (OCR) images from scanned documents. There is a large research literature \cite{chen2021text} addressing these present challenges to obtain the best STR model predictor.

Many of these works, however, only focus on solving the main problem of increasing model accuracy. They do not address the problem of explaining why STR models work in general. Previous experiments focus on attention maps \cite{wu2018scan, shi2018aster, siddiqui2020using} that only provide a simple analysis and do not focus on providing explainability to STR. The vast majority of XAI methods out there \cite{miller2019explanation,burkart2021survey,zhou2021evaluating} in computer vision are usually only evaluated on single-class image classification tasks and are generally ineffective in explaining STR networks. STR predictors are black-box models constituting various deep neural network architectures and comprising of a multi-class output prediction, making them more challenging to understand. Motivated by this, a study is presented merging both XAI and STR, called Scene Text Recognition (STR) Explainability. We ask the question related to XAI, "Why do STR models work?" \cite{gilpin2018explaining,miller2019explanation,molnar2018guide,shickel2020sequential}.
This question focuses on trying to explain the input data of a deep learning model and persuades us to learn more about the need to explain why STR models work in general. Not only does explainability benefit AI engineers, but they can also provide the explanations to convince non-AI experts that these models are trustable and safe to use, even in high stake decisions.

\begin{figure}[!t]
  \includegraphics[width=\columnwidth]{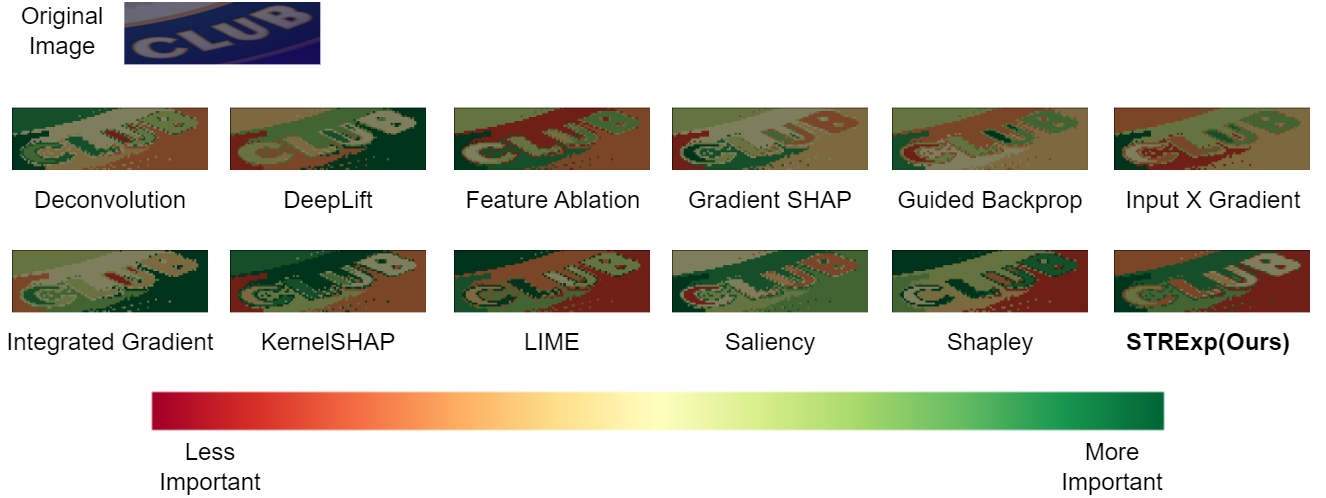}
  \caption[STRExp Sample Data Explanations]{Examples of attribution-based method explanations, wherein greener areas are more important and red areas are less important. When executed on STR models, previous attribution-based methods produce inconsistent and ineffective data explanations (first 11 images), that produce false positives (green areas far away from the text), and false negatives (red areas near the text). STRExp reduces this explanation inconsistency (12th image) by placing more importance in the actual text areas of the image.}
  \label{fig:sample_dataexp}
\end{figure}

In this work, the problem of interest is in STR Data Explainability, which focuses on providing explainability to the input data. To the best of our knowledge, there is only little work on Explainable STR. Thus, we focus on recent literatures, called interpretable multi-label classification, that have some similar characteristics with our problem. However, integrating these to previous attribution-based methods lead to inconsistent and ineffective explanations in STR models (Fig. \ref{fig:sample_dataexp}). This is because these works only try to explain multi-label class models in the global context \cite{luaces2012binary,kokhlikyan2020captum,ribeiro2016should} (Fig. \ref{fig:strexp}). To solve this problem, we propose to execute the explanations to include the local individual character explanations. Combining both the local and global explanations produces stronger average explanations on the input. Data explanation provides evidence to understand why an STR model works. This increases model trustability and simplicity \cite{zhou2021evaluating}.

In summary, the contributions of this work are as follows: (1) To the best of our knowledge, this work is the first in creating a new data explainability framework specifically made for the task of STR. (2) The local explanations of STR models are leveraged and combined with the global explanations to reduce the inconsistency and ineffectiveness of previous attribution-based methods. We call this STRExp. (3) Our method is then benchmarked across different STR datasets, and show superior explanation performance when compared to previous attribution-based methods across different STR models.

\section{Related Works}
The area relating to the task of explaining the input features of a host model are typically associated with attribution-based explanations. In the current literature, attribution-based explanation methods \cite{zhou2021evaluating,lundberg2017unified,ribeiro2016should} make up the majority of machine learning explanations. These methods normally interpret which of the individual input features contribute most to the output's prediction. However, most of their explanation examples in computer vision are applied to single-label image classification problems. Existing works implementing them into multi-label classification systems show simple and ineffective explanations. Thus, the task of integrating them into Scene Text Recognition (STR), a subset of a multi-label classification problem, is a major challenge in itself.

Attribution-based explanations leverage the model features extracted in order to create an interpretable data representation. Popular methods such as LIME \cite{ribeiro2016should} and SHAP \cite{lundberg2017unified} provide feature-based explanations in terms of highlighting pixels/superpixels. However, in the context of a multi-label classification problem like STR, these attribution-based methods suggest on transforming the former into a single-label class output \cite{luaces2012binary,ribeiro2016should}. Thus, this form of explainability execution can only output the model's attributions in the global context. In our case, we also take into consideration the individual characters of the STR model to reduce the  ineffectiveness produced by these previous method's explanations.

\section{Methodology}
\subsection{STR Attribution-based methods}
An attribution-based method, $\mathbb{A}$, consists of an iterative algorithm that uses either the forwardpass parameters, backwardpass parameters, or both, to compute an explanation (Fig. \ref{fig:sample_dataexp}) of a model. Given an STR model $M$, its parameters $\theta$, and one of the attribution-based methods $A_{i} \in \mathbb{A}$, the latter's explanation is given by $E(X) = p^{attr} = A_{i}(M_{\theta}(X))$. This is also called the attributions of $A_{i}$. Thus, for each input feature $X = \{X_{1}, X_{2},..., X_{n}\}$, its attribution representation is also given by $E(X) = \{E(X_1),...,E(X_n)\}$.

\begin{figure}[!t]
  \includegraphics[width=\columnwidth]{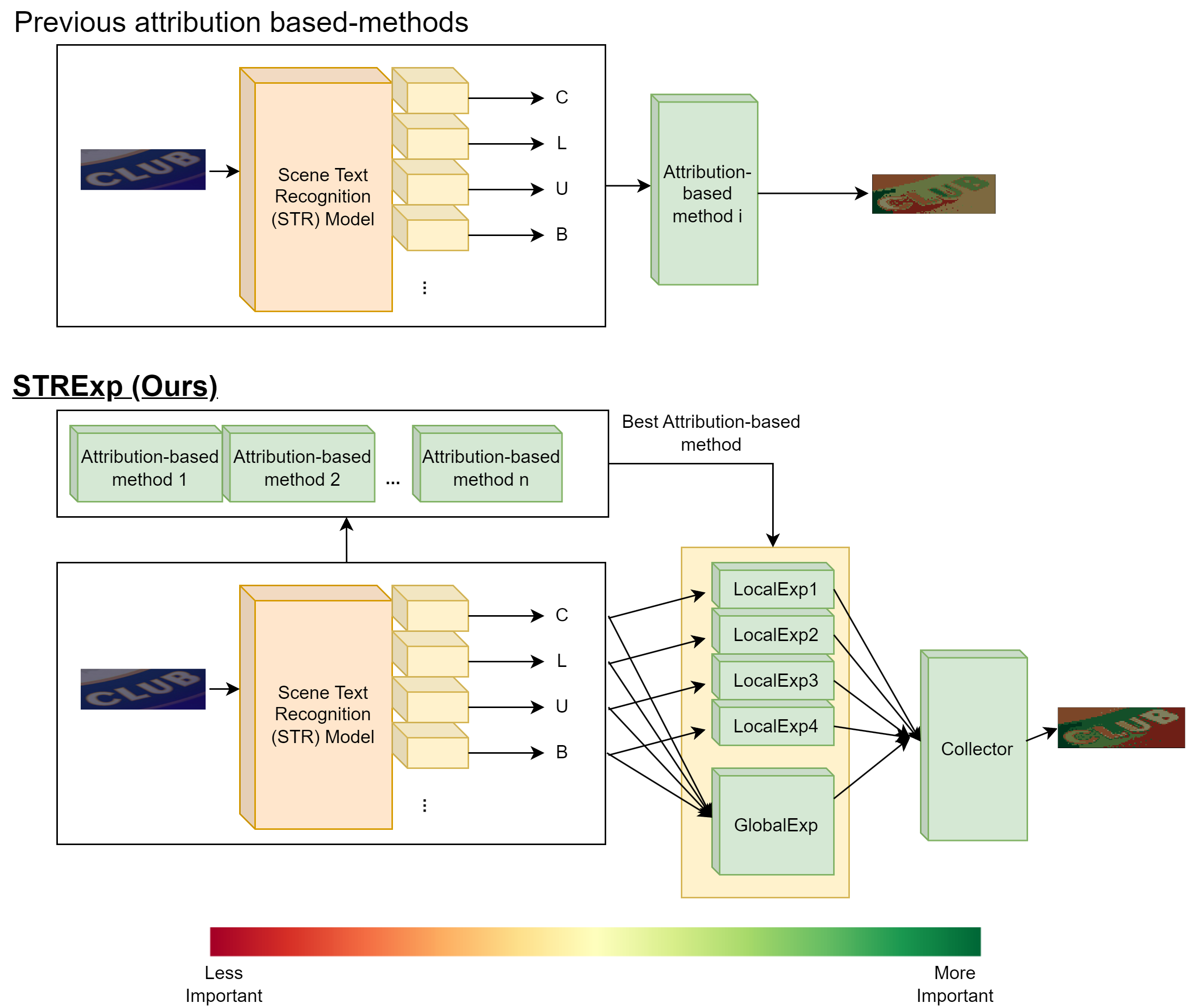}
  \caption[STRExp Diagram]{Previous attribution-based methods only execute their explanations in the global context of the STR model. We further improve data explainability by querying the best attribution-based methods, and then combining both the local and global explanations. This new method is called STRExp.}
  \label{fig:strexp}
\end{figure}

The calculation of the attribution $E(X)$ is done using either the forwardpass parameters $L$, backwardpass parameters $G$, or both, depending on the attribution-based method. The forwardpass parameters refers to the layer maps during model prediction, $L=\{L_{1}, L_{2},..., L_{n}\}$, where $L_{1}$ refers to the first layer, $L_{2}$ refers to the second layer, etc. The backwardpass parameters refers to the gradients of each layer during backpropagation, set to $G=\frac{\partial e(Y,\hat{Y})}{\partial\theta}=\{\frac{\partial e(Y,\hat{Y})}{\partial\theta_{L_{1}}}, \frac{\partial e(Y,\hat{Y})}{\partial\theta_{L_{2}}},..., \frac{\partial e(Y,\hat{Y})}{\partial\theta_{L_{n}}}\}$, where $e(Y,\hat{Y})$ is some error function between the target value $Y$ and the predicted value $\hat{Y}$ with respect to the model parameters at a specific layer $\theta_{L}$.

\subsection{Global and Local Explanations}
For each input features $X = \{X_{1}, X_{2},..., X_{n}\}$, an STR model $M$ with parameters $\theta$ must predict a sequence of label space $Y = \{Y_{1},Y_{2},..., Y_{n}\}$. Its corresponding score is given by $r = R(M_{\theta}(x))$, where $R$ is some function that calculates the mean output to convert it into a single-label space. Each attribution-based method is set as $A_{i} \in \mathbb{A}$. Thus, each attribution/explanation output is then set to $p^{attr}_{global} = A_{i}(x, r)$, which indicates that this explanation is in the global context of the model $M_{\theta}$, capturing the STR model's distribution $Y = \{Y_{1},Y_{2},..., Y_{n}\} =  P(\{X_{1},X_{2}, ..., X_{n}\}|\theta)$. After evaluating each explanations, an explainability evaluator metric $\gamma$ is used to query the best $A_i$. We used selectivity \cite{montavon2018methods} for $\gamma$. For each input feature segmentation, $X = \{X_{1}, X_{2},..., X_{n}\}$, its corresponding attribution, $s^{attr}_{x} = \mathbb{E}(p^{attr}_{x})$, is acquired. Finally, the total performance for $A_{i}$ is set to $z_{i} = \gamma([s^{attr}_{x_1}, s^{attr}_{x_2},..., s^{attr}_{x_m}])$ for all total segmentations $m$. After minimizing the Selectivity Area Under the Curve (AUC), $min(E_{A_{1}}(X)), E_{A_{2}}(X)), ..., E_{A_{t}}(X)))$, the best attribution-based method $B = \mathbb{A}(z)$ is queried.

\begin{figure}
  \includegraphics[width=\columnwidth]{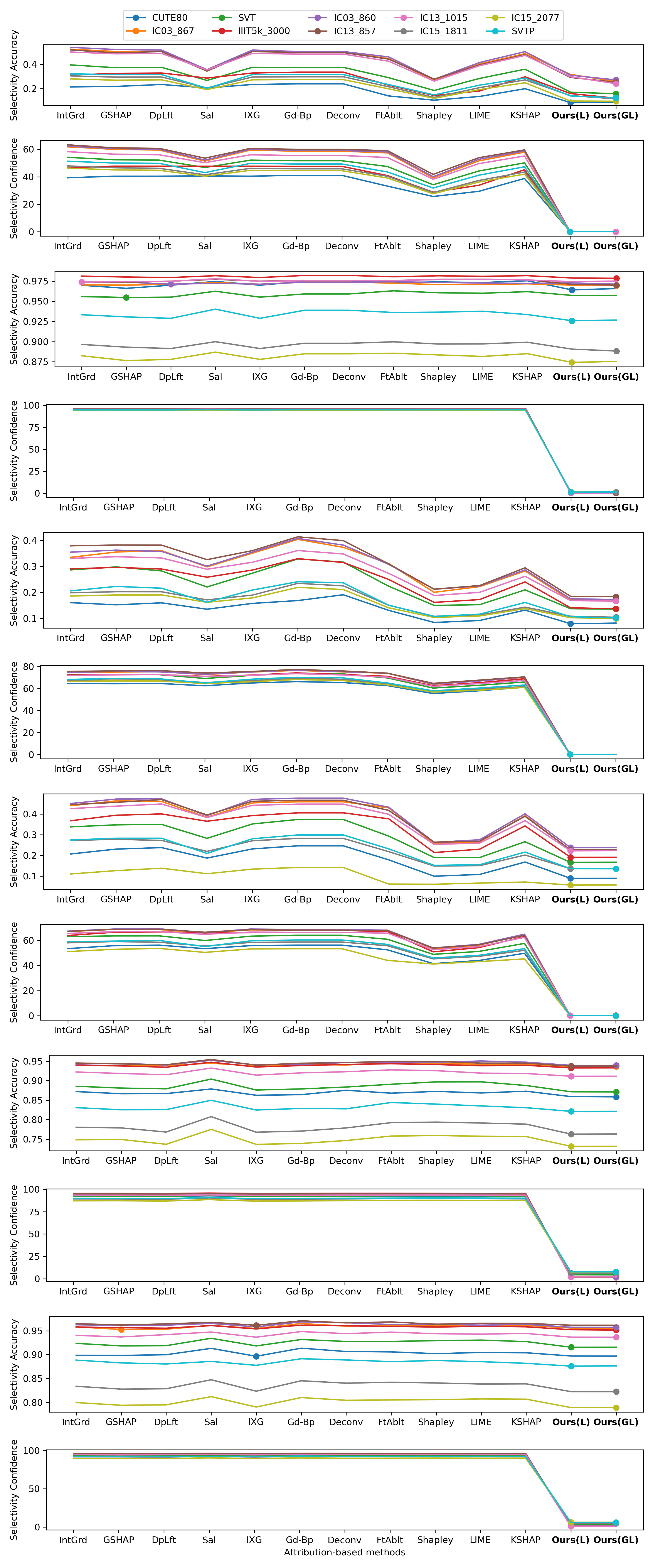}
  \caption[VITSTR, PARSeq, TRBA, SRN, ABINET, MATRN Selectivity Quantitative]{From top to bottom: VITSTR\cite{atienza2021vision}(1st \& 2nd figure), PARSeq\cite{bautista2022scene}(3rd \& 4th figure), TRBA\cite{baek2019wrong}(5th \& 6th figure), SRN\cite{yu2020towards}(7th \& 8th figure), ABINET\cite{fang2021read}(9th \& 10th figure) and MATRN\cite{na2022multi}(11th \& 12th figure) quantitative results.}
  \label{fig:attr_quantitative_vitstr_parseq_trba_srn_abinet_matrn_acc}
\end{figure}

The problem with using the global context is that it provides incoherent gradients to the attribution-based method. You can only impose a single target value of $\frac{\partial e(Y=1,\hat{Y})}{\partial\theta}$ during backpropagation for all STR outputs. This leads to the main problem of producing those inconsistent and ineffective data explanations. To solve this problem, we propose to involve the local explanations of each individual character prediction. Thus, the value of its gradients will be $G_{c}=\{\frac{\partial e(Y=c_{1},\hat{Y})}{\partial\theta}, \frac{\partial e(Y=c_{2},\hat{Y})}{\partial\theta},..., \frac{\partial e(Y=c_{n},\hat{Y})}{\partial\theta}\}$ where $c_{i}$ are the individual target character values of the STR model. This solves the gradient incoherency produced by the global explanations. We build from the best attribution-based method $B$ and execute it in the local context of the STR model. For each image $I$ with input features $X = \{X_{1}, X_{2},..., X_{n}\}$, there is a fixed number of total characters $c \in C$. The local score of the individual character is then computed by $r_{c_{k}} = R(M_{\theta}(x), k)$ and its attribution is computed by $p^{attr}_{local} = B(x, r_{c_{k}})$. Here, $c_{k}$ is the character $c$ at the $kth$ position. This attribution describes the STR model's single-character prediction distribution $Y_{i} =  P(\{X_{1},X_{2}, ..., X_{n}\}|\theta)$, without depending on the outputs of the other characters. This process is executed for each other characters $c$ until there are $k$ individual character attributions. As described in Fig. \ref{fig:strexp}, these local explanations are also combined with the global explanation of the best attribution-based method, $p^{attr}_{global} = B(x, r)$ to further improve the data explanation performance. The final attribution is then set to $p^{final}_{final} = \mathbb{E}(p^{attr}_{local}, p^{attr}_{global})$, and then its selectivity is again computed to compare it with previous attribution-based methods. The code can be found in \url{https://github.com/markytools/strexp}.

\section{Results and Discussion}
\subsection{Metric Evaluation}

The metrics used are derived from the selectivity metric. Explanation selectivity \cite{montavon2018methods} measures how fast a function $f(x)$ goes down when removing features with the highest relevance. The highest scored segmentation feature from an attribution-based method explanation is removed and the STR performance is plotted (y-axis). This process is repeated iteratively until all input features have been removed (x-axis). The final selectivity value will be the area under the curve (AUC). The best attribution-based method will have the lowest selectivity (i.e. the lowest AUC), suggesting that removing this method’s most important areas in the image first will coincide with a larger drop in the STR model performance compared to other attribution-based methods. Two STR performances are used to derive two new metrics from selectivity. One is using the STR text accuracy called Selectivity Accuracy, while the other is using the STR mean confidence called Selectivity Confidence. For each dataset, the average selectivity is computed across all images.

 \begin{figure}[!t]
  \includegraphics[width=\columnwidth]{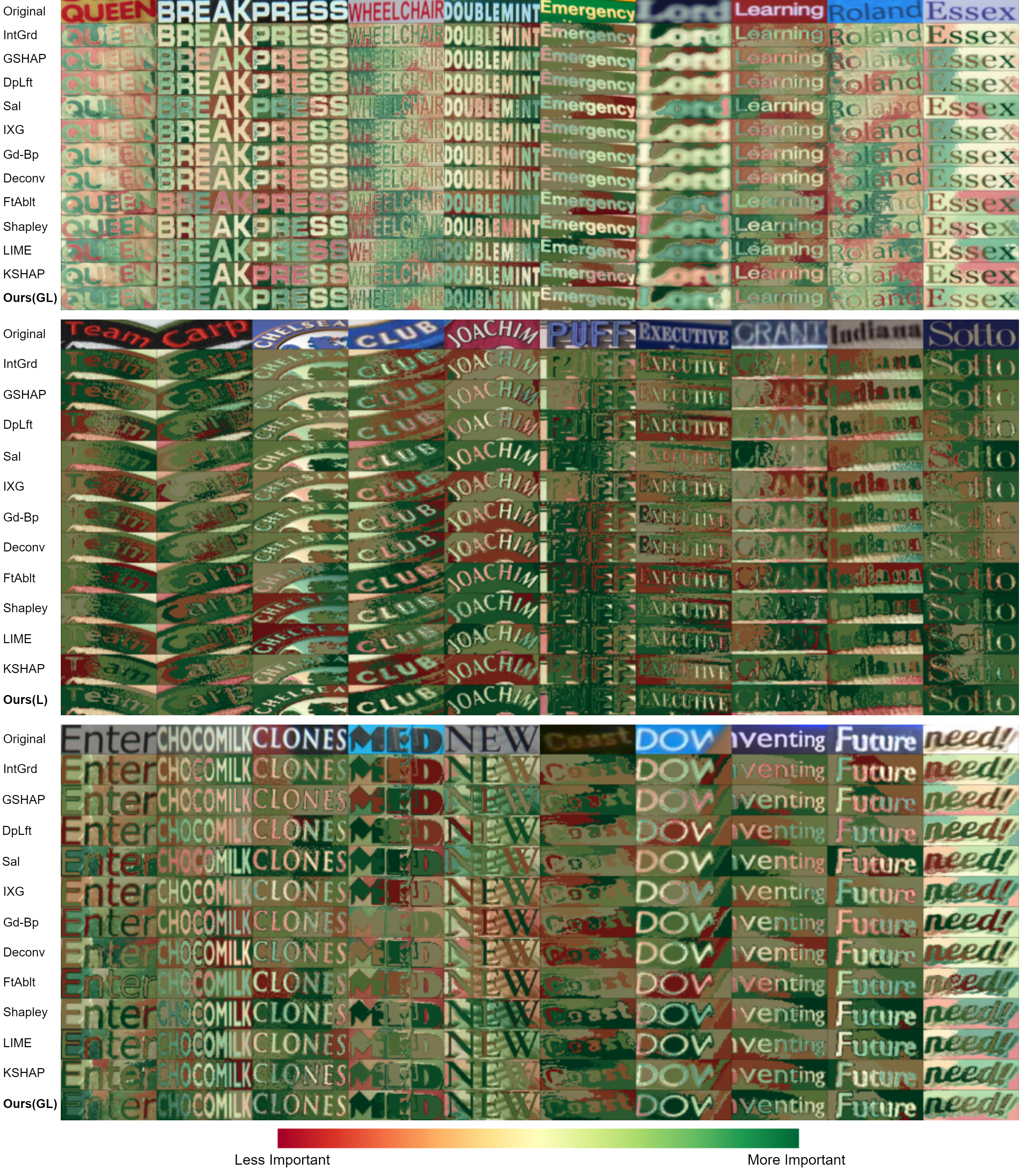}
  \caption[PARSeq, SRN, and TRBA Qualitative]{From top to bottom: PARSeq\cite{bautista2022scene}(1st \& 2nd figure), SRN\cite{yu2020towards}(3rd \& 4th figure), and TRBA\cite{baek2019wrong}(5th \& 6th figure) qualitative results.}
  \label{fig:parseq_srn_trba_qualitative}
\end{figure}

\subsection{STRExp Quantitative Results}
 Using the selectivity metric, the quantitative results are obtained to compare STRExp with previous attribution-based methods across different STR real datasets and across different STR models.
 For evaluation, the STR model architectures used are: VITSTR \cite{atienza2021vision}, TRBA \cite{baek2019wrong}, PARSeq \cite{bautista2022scene}, SRN \cite{yu2020towards}, ABINET \cite{fang2021read}, and MATRN \cite{na2022multi}. The attribution-based methods (and their abbreviations) to be compared are \cite{kokhlikyan2020captum}: Integrated Gradients (IntGrd), GradientSHAP (GSHAP), DeepLift (DpLft), Saliency (Sal), Input X Gradient (IXG), Guided Backprop (Gd-Bp), Deconvolution (Deconv), KernelSHAP (KSHAP), Feature Ablation (FtAblt), LIME (LIME), and Shapley (Shapley). The evaluation is done on different real-world STR test datasets \cite{chen2021text}: CUTE80, SVT, IIIT5k\textunderscore3000, SVTP, IC03\textunderscore860, IC03\textunderscore867, IC13\textunderscore857, IC13\textunderscore1015, IC15\textunderscore1811, IC15\textunderscore2077.
 
 Fig. \ref{fig:attr_quantitative_vitstr_parseq_trba_srn_abinet_matrn_acc} shows the benchmarks of STRExp. The y-axis in the figure represents either the selectivity accuracy or selectivity confidence metric. The x-axis represents the abbreviations of the attribution-based methods to be compared. The line colors represent the different STR datasets. Thus, a single (x,y) coordinate represents the selectivity of the attribution-based method when evaluated on that dataset. The dot/point of each line signifies the attribution-based method that has the lowest selectivity on that dataset. Ours(GL) refers to STRExp having its output combined with both the global and local explanations, while Ours(L) refers to STRExp only using the local explanations. STRExp is executed to produce the explanations of the VITSTR STR Model \cite{atienza2021vision}, PARSeq STR Model \cite{bautista2022scene}, TRBA STR Model\cite{baek2019wrong}, SRN STR Model\cite{yu2020towards}, ABINET STR Model \cite{fang2021read}, and MATRN STR Model \cite{na2022multi}. The results show that our proposed method, STRExp, generally has the lowest selectivity accuracy and lowest selectivity confidence in the majority of cases compared to previous attribution-based methods evaluated across different STR datasets.

\subsection{STRExp Qualitative Results}
The general traits/conditions for a good qualitative image are: (1) The best attribution-based method will have greener areas inside and near the text compared to previous methods, because hiding these areas of the image first (according to the selectivity metric) will produce a greater accuracy drop compared to other areas. (2) The best attribution-based method will have less red areas inside and near the text compared to previous methods. It does not make any sense to hide the text areas last when trying to produce a lower selectivity.

In the top of Fig. \ref{fig:parseq_srn_trba_qualitative}, STRExp is evaluated on PARSeq\cite{bautista2022scene} STR Model in some image samples of the IC03\textunderscore867 (first 5 images) and the IC13\textunderscore857 (last 5 images) datasets, showing how a low selectivity accuracy from Fig. \ref{fig:attr_quantitative_vitstr_parseq_trba_srn_abinet_matrn_acc} impacts the results visually. 
The first image column with the "QUEEN" text has more greener colors near and inside the text in STRExp compared to previous attribution-based methods. In the second column "BREAK", STRExp has more greener areas and less redder areas in the text compared to previous methods. This trend follows for all other image columns in the figure.

In the middle of Fig. \ref{fig:parseq_srn_trba_qualitative}, another STR Model, SRN\cite{yu2020towards}, is evaluated on CUTE80 (first 5 images) and SVT (last 5 images) datasets, showing how a low selectivity accuracy from Fig. \ref{fig:attr_quantitative_vitstr_parseq_trba_srn_abinet_matrn_acc} impacts the results visually. In the bottom of Fig. \ref{fig:parseq_srn_trba_qualitative}, the TRBA\cite{baek2019wrong} STR Model is evaluated on the IC03\textunderscore860 (first 5 images) and IC15\textunderscore2077 (last 5 images) datasets, showing how a low selectivity accuracy from Fig. \ref{fig:attr_quantitative_vitstr_parseq_trba_srn_abinet_matrn_acc} impacts the results visually.

\section{Conclusion}
This work proposes STRExp that leverages the local individual character explanations to produce better STR explanations compared to previous attribution-based methods.

\section{Acknowledgement}
DOST ERDT scholarships and ERDT FRDG.

\bibliographystyle{IEEEbib}
\bibliography{refs}

\begin{thebibliography}{10}

\bibitem{baek2019wrong}
Jeonghun Baek, Geewook Kim, Junyeop Lee, Sungrae Park, Dongyoon Han, Sangdoo Yun, Seong~Joon Oh, and Hwalsuk Lee,
\newblock ``What is wrong with scene text recognition model comparisons? dataset and model analysis,''
\newblock in {\em Proceedings of the IEEE/CVF International Conference on Computer Vision}, 2019, pp. 4715--4723.

\bibitem{yu2020towards}
Deli Yu, Xuan Li, Chengquan Zhang, Tao Liu, Junyu Han, Jingtuo Liu, and Errui Ding,
\newblock ``Towards accurate scene text recognition with semantic reasoning networks,''
\newblock in {\em Proceedings of the IEEE/CVF Conference on Computer Vision and Pattern Recognition}, 2020, pp. 12113--12122.

\bibitem{chen2021text}
Xiaoxue Chen, Lianwen Jin, Yuanzhi Zhu, Canjie Luo, and Tianwei Wang,
\newblock ``Text recognition in the wild: A survey,''
\newblock {\em ACM Computing Surveys (CSUR)}, vol. 54, no. 2, pp. 1--35, 2021.

\bibitem{wu2018scan}
Yi-Chao Wu, Fei Yin, Xu-Yao Zhang, Li~Liu, and Cheng-Lin Liu,
\newblock ``Scan: Sliding convolutional attention network for scene text recognition,''
\newblock {\em arXiv preprint arXiv:1806.00578}, 2018.

\bibitem{shi2018aster}
Baoguang Shi, Mingkun Yang, Xinggang Wang, Pengyuan Lyu, Cong Yao, and Xiang Bai,
\newblock ``Aster: An attentional scene text recognizer with flexible rectification,''
\newblock {\em IEEE transactions on pattern analysis and machine intelligence}, vol. 41, no. 9, pp. 2035--2048, 2018.

\bibitem{siddiqui2020using}
Sahar Siddiqui, Elena Sizikova, Gemma Roig, Najib~J Majaj, and Denis~G Pelli,
\newblock ``Using human psychophysics to evaluate generalization in scene text recognition models,''
\newblock {\em arXiv preprint arXiv:2007.00083}, 2020.

\bibitem{miller2019explanation}
Tim Miller,
\newblock ``Explanation in artificial intelligence: Insights from the social sciences,''
\newblock {\em Artificial intelligence}, vol. 267, pp. 1--38, 2019.

\bibitem{burkart2021survey}
Nadia Burkart and Marco~F Huber,
\newblock ``A survey on the explainability of supervised machine learning,''
\newblock {\em Journal of Artificial Intelligence Research}, vol. 70, pp. 245--317, 2021.

\bibitem{zhou2021evaluating}
Jianlong Zhou, Amir~H Gandomi, Fang Chen, and Andreas Holzinger,
\newblock ``Evaluating the quality of machine learning explanations: A survey on methods and metrics,''
\newblock {\em Electronics}, vol. 10, no. 5, pp. 593, 2021.

\bibitem{gilpin2018explaining}
Leilani~H Gilpin, David Bau, Ben~Z Yuan, Ayesha Bajwa, Michael Specter, and Lalana Kagal,
\newblock ``Explaining explanations: An overview of interpretability of machine learning,''
\newblock in {\em 2018 IEEE 5th International Conference on data science and advanced analytics (DSAA)}. IEEE, 2018, pp. 80--89.

\bibitem{molnar2018guide}
Christoph Molnar,
\newblock ``A guide for making black box models explainable,''
\newblock {\em URL: https://christophm. github. io/interpretable-ml-book}, 2018.

\bibitem{shickel2020sequential}
Benjamin Shickel and Parisa Rashidi,
\newblock ``Sequential interpretability: Methods, applications, and future direction for understanding deep learning models in the context of sequential data,''
\newblock {\em arXiv preprint arXiv:2004.12524}, 2020.

\bibitem{luaces2012binary}
Oscar Luaces, Jorge D{\'\i}ez, Jos{\'e} Barranquero, Juan~Jos{\'e} del Coz, and Antonio Bahamonde,
\newblock ``Binary relevance efficacy for multilabel classification,''
\newblock {\em Progress in Artificial Intelligence}, vol. 1, no. 4, pp. 303--313, 2012.

\bibitem{kokhlikyan2020captum}
Narine Kokhlikyan, Vivek Miglani, Miguel Martin, Edward Wang, Bilal Alsallakh, Jonathan Reynolds, Alexander Melnikov, Natalia Kliushkina, Carlos Araya, Siqi Yan, et~al.,
\newblock ``Captum: A unified and generic model interpretability library for pytorch,''
\newblock {\em arXiv preprint arXiv:2009.07896}, 2020.

\bibitem{ribeiro2016should}
Marco~Tulio Ribeiro, Sameer Singh, and Carlos Guestrin,
\newblock ``" why should i trust you?" explaining the predictions of any classifier,''
\newblock in {\em Proceedings of the 22nd ACM SIGKDD international conference on knowledge discovery and data mining}, 2016, pp. 1135--1144.

\bibitem{lundberg2017unified}
Scott~M Lundberg and Su-In Lee,
\newblock ``A unified approach to interpreting model predictions,''
\newblock {\em Advances in neural information processing systems}, vol. 30, 2017.

\bibitem{montavon2018methods}
Gr{\'e}goire Montavon, Wojciech Samek, and Klaus-Robert M{\"u}ller,
\newblock ``Methods for interpreting and understanding deep neural networks,''
\newblock {\em Digital Signal Processing}, vol. 73, pp. 1--15, 2018.

\bibitem{atienza2021vision}
Rowel Atienza,
\newblock ``Vision transformer for fast and efficient scene text recognition,''
\newblock in {\em International Conference on Document Analysis and Recognition}. Springer, 2021, pp. 319--334.

\bibitem{bautista2022scene}
Darwin Bautista and Rowel Atienza,
\newblock ``Scene text recognition with permuted autoregressive sequence models,''
\newblock in {\em European Conference on Computer Vision}. Springer, 2022, pp. 178--196.

\bibitem{fang2021read}
Shancheng Fang, Hongtao Xie, Yuxin Wang, Zhendong Mao, and Yongdong Zhang,
\newblock ``Read like humans: Autonomous, bidirectional and iterative language modeling for scene text recognition,''
\newblock in {\em Proceedings of the IEEE/CVF Conference on Computer Vision and Pattern Recognition}, 2021, pp. 7098--7107.

\bibitem{na2022multi}
Byeonghu Na, Yoonsik Kim, and Sungrae Park,
\newblock ``Multi-modal text recognition networks: Interactive enhancements between visual and semantic features,''
\newblock in {\em European Conference on Computer Vision}. Springer, 2022, pp. 446--463.

\end{thebibliography}

\end{document}